\documentclass[10pt,twocolumn,letterpaper]{article}

\usepackage{cvpr}
\usepackage{times}
\usepackage{epsfig}
\usepackage{graphicx}
\usepackage{amsmath}
\usepackage{amssymb}

\usepackage{amsmath}
\usepackage{array}
\usepackage{subfigure}

\usepackage{balance}

\usepackage{multirow}
\usepackage{booktabs}

\usepackage{amssymb}

\usepackage{xspace}
\makeatletter
\DeclareRobustCommand\onedot{\futurelet\@let@token\@onedot}
\def\@onedot{\ifx\@let@token.\else.\null\fi\xspace}
\def\eg{\emph{e.g}\onedot} 
\def\ie{\emph{i.e}\onedot}

\makeatother

\usepackage{pifont}%
\newcommand{\cmark}{\ding{51}}%
\newcommand{\xmark}{\ding{55}}%

\usepackage{authblk}

\usepackage[pagebackref=true,breaklinks=true,letterpaper=true,colorlinks,bookmarks=false]{hyperref}

 \cvprfinalcopy %

\def\cvprPaperID{1648} %

\begin{document} 
	
	\title{Normalized and Geometry-Aware Self-Attention Network \\ for Image Captioning}

\author[$^{1,4}$ ]{
	Longteng Guo
}
\author[$^{1}$]{
	\; Jing Liu%
}
\author[$^{1}$]{
	\; Xinxin Zhu
}
\author[$^{2}$]{
	\; Peng Yao
}
\author[$^{3}$]{
	\; Shichen Lu
}
\author[$^{1}$]{
	\;Hanqing Lu
}

\affil[ ]{$^1$National Laboratory of Pattern Recognition, Institute of Automation, Chinese Academy of Sciences\; $^2$University of Science and Technology Beijing \; $^3$Wuhan University\;  }
\affil[4]{University of Chinese Academy of Sciences}
\affil[ ]{ 
	\tt\small \{longteng.guo,jliu,xinxin.zhu,luhq\}@nlpr.ia.ac.cn,S20180598@xs.ustb.edu.cn,sclu@whu.edu.cn}

\renewcommand\Authsep{ } 
\renewcommand\Authands{ }
	
	\maketitle
	
	\begin{abstract}
		Self-attention (SA) network has shown profound value in image captioning. In this paper, we improve SA from two aspects to promote the performance of image captioning. First, we propose Normalized Self-Attention (NSA), a reparameterization of SA that brings the benefits of normalization inside SA. While normalization is previously only applied outside SA, we introduce a novel normalization method and demonstrate that it is both possible and beneficial to perform it on the hidden activations inside SA. Second, to compensate for the major limit of Transformer that it fails to model the geometry structure of the input objects, we propose a class of Geometry-aware Self-Attention (GSA) that extends SA to explicitly and efficiently consider the relative geometry relations between the objects in the image. To construct our image captioning model, we combine the two modules and apply it to the vanilla self-attention network. We extensively evaluate our proposals on MS-COCO image captioning dataset and superior results are achieved when comparing to state-of-the-art approaches. Further experiments on three challenging tasks, \ie video captioning, machine translation, and visual question answering, show the generality of our methods. 
	\end{abstract}

	\section{Introduction} 
	Automatically generating captions for images, namely image captioning \cite{kiros2014multimodal,xu2015show}, 
	has emerged as a prominent research problem at the intersection of computer vision (CV) and natural language processing (NLP). 
	This task is challenging as it requires to first recognize the objects in the image, the relationships between them, 
	and finally properly organize and describe them in natural language.

	Inspired by the sequence-to-sequence model for machine translation, 
	most image captioning approaches adopt an encoder-decoder paradigm, which 
	uses a deep convolutional neural network (CNN) to encode the input image as a vectorial representation,  
	and a recurrent neural network (RNN) based caption decoder to generate the output caption. 
	Recently, \textit{self-attention} (SA) networks, denoted as SANs, have been introduced by \cite{zhu2018captioning,yu2019multimodal} to replace conventional RNNs in image captioning. 
	Since its first introduction in Transformer \cite{vaswani2017attention}, 
	SA and its variants have shown promising empirical results in a wide range of CV \cite{zhang2018self,hu2018relation,wang2018non,fu2019dual,liu2019large,fu2020contextual} and NLP \cite{tan2018deep,devlin2018bert,yang2019context} tasks. 
	Although SAN-based framework has achieved state-of-the-art performance in image captioning, it remains two problems to be solved.

	Firstly, SA is susceptible to the internal covariate shift \cite{Ioffe2015Batch} problem. 
	Typically, SA is regarded as a mapping of a set of query and key/value pairs. 
	We observe, from another perspective, that computation of the attention weights in SA could be 
	considered as feeding the queries into a fully-connected layer, 
	whose parameters are dynamically computed according to the inputs.   
	Problem could happen when the distribution of the queries shifts due to the change in network parameters during training.  
	That is, the subsequent layers have to continuously adapt to the new input distribution, and consequently, SA may not be learned effectively. 
	This problem is called ``Internal Covariate Shift" in \cite{Ioffe2015Batch} –--- 
	the tendency that the distribution of activations drifts during training in a feed-forward network. 

	To eliminate the internal covariate shift problem inside SA, in this paper, we introduce an effective reparameterization of SA, named Normalized Self-Attention (NSA). 
	NSA performs a novel normalization method on the hidden activations of SA to fix their distributions. 
	By doing so, we can effectively decouple the fully-connected layer's parameters from those of other layers, 
	leading to a better-conditioned optimization of SA. 
	While Layer Normalization (LN) \cite{ba2016layer} is proven to be very critical for enabling the convergence of Transformer, 
	however, LN is only applied \textit{outside} SA blocks. 
	To our knowledge, there has not been any deep exploration to find a suitable normalization method \textit{inside} SA.  
	We demonstrate that our NSA can collaborate with LN to bring improved generalization for SA-based networks.

	Another critical issue in SA is its inability to model the geometric relationships among input elements. 
	The vanilla self-attention treats its inputs as ``\textit{bag-of-features}", 
	simply neglecting their structure and the relationships between them. 
	However, the objects in the image, from which the region-based visual features are extracted for image captioning, 
	inherently have geometric structure --- 2D spatial layout and variations in scale/aspect ratio. 
	Such inherent geometric relationships between objects play a very complex yet critical role in understanding the image content. 
	One common solution to inject position information into SA is adding representations of absolute positions to each element of the inputs, as is often used in the case of 1D sentences.  
	Nonetheless, this solution does not work well for image captioning because 
	the 2D geometry relations between objects are harder to infer from their absolute positions.
	
	We present a more efficient approach to the above problem: explicitly incorporating \textit{relative} geometry relationships between objects into SA. 
	The module is named Geometry-aware Self-Attention (GSA). 
	GSA extends the original attention weight into two components: the original content-based weight, 
	and a new geometric bias, which is efficiently calculated by the relative geometry relations and, importantly, the \textit{content} of the associated elements, \ie query or key. 
	
	By combining both NSA and GSA, we obtain an enhanced SA module. 
	We then construct our Normalized and Geometry-aware Self-Attention Network, namely NG-SAN, by replacing 
	the vanilla SA modules in the encoder of the self-attention network with the proposed one. 
	Extensive experiments on MS-COCO validates the effectiveness of our proposals. 
	In particular, our NG-SAN establishes a new state-of-the-art on the MS-COCO evaluation sever, 
	improving the best single-model result in terms of CIDEr from 125.5 to 128.6. 
	To demonstrate the generality of NSA, we further present video captioning, machine translation, and visual question answering experiments on the VATEX, WMT 2014 English-to-German, and VQA-v2 datasets, respectively. 
	On top of the strong Transformer-based baselines, our methods can consistently increase accuracies on all tasks at a negligible extra computational cost.
	
	To summarize, the main contributions of this paper are three-fold: 
	
	\begin{itemize}
		\item We presented Normalized Self-Attention, an effective reparameterization of self-attention, 
		which brings the benefits of normalization technique inside SA. 
		\item We introduce a class of Geometry-aware Self-Attention that explicitly makes use of the 
		relative geometry relationships and the content of objects to aid image understanding. 
		\item By combining the two modules and apply it on the self-attention network, 
		we establish a new state-of-the-art on the MS-COCO image captioning benchmark. 
		Further experiments on video captioning, machine translation, and visual question answering tasks demonstrate the generality of our methods.

	\end{itemize}

	\section{Related Work}

	\subsection{Image Captioning}
	Existing image captioning approaches typically follows the CNN-RNN architecture \cite{vinyals2017show}.  
	Recently, a variety of improving works have been proposed. 
	\cite{xu2015show} introduces soft and hard attention mechanisms to automatically focus on salient objects when generating each word. 
	\cite{guo2019show} mimics human polishing process with a ruminant decoder. 
	\cite{anderson2017bottom} uses an object detector to propose salient image regions (objects) and extract for each object a feature vector, 
	which are then used as inputs for attention mechanism.  
	\cite{Rennie2016Self} introduces reinforcement-learning with a self-critical reward for model training. 
	Recently, \cite{zhu2018captioning} and \cite{yu2019multimodal} propose to 
	replace conventional RNN with the Transformer architecture, achieving state-of-the-art performance.  
	However, more deep exploration of the self-attention module in Transformer is not conducted on the task of image captioning, 
	which motivates our work in this paper.

	\subsection{Normalization} 
	Normalization \cite{Ioffe2015Batch} has become a critical ingredient in constructing a deep neural network. 
	It is proposed by Batch normalization (BN) \cite{Ioffe2015Batch} to control the distributions of the internal activations of feed-forward neural networks, thereby reducing internal covariate shift. 
	Several variants of normalization method such as Layer Normalization (LN) \cite{ba2016layer}, Instance Normalization (IN) \cite{ulyanov2016instance}, 
	and Group Normalization \cite{wu2018group} have been developed mainly to reduce the mini-batch dependencies inherent in BN. 
	LN operates along the channel dimension for each individual element in an example.
	IN performs BN-like computation but only for each sample. %
	Though BN and LN have been adopted in networks that contain the SA module, \eg Transformer, 
	they are typically used outside the SA module. 
	For the first time, our normalized self-attention brings the benefit of normalization inside the SA module. 
	
	\subsection{Position encoding in self-attention networks} 
	To inject sequence ordering into SA module, in Transformer, 
	absolute position encodings based on sinusoids are added to the input elements both in the encoder and decoder.
	Recently, \cite{shaw2018self} modulates SA by incorporating the relative distances between sequence elements. 
	\cite{hu2018relation} proposes an SA-like module for object detection, which multiplies a new relation weight on the original self-attention weight, 
	and is used by \cite{herdade2019image} in Transformer. 
	Its relation weight is computed solely with the relative coordinates and sizes between bounding boxes.  
	Different from these works, our GSA module explores a broader range of geometric biases 
	that involve not only the geometry information but also the content of the associated objects.

	\section{Preliminaries}
	\subsection{Self-Attention (SA)}
	We first review a basic form of self-attention, called ``Scaled Dot-Product Attention", which is first proposed as a core component in Transformer.
	
	The self-attention layer first transforms a set of $N$ $d_k$-dimensional vectors, packed into a matrix $X \in \mathbb{R}^{N \times d_k}$, into queries $Q \in \mathbb{R}^{N \times d}$, keys $K \in \mathbb{R}^{N \times d}$, and values $V\in \mathbb{R}^{N \times d}$ given by 
	$Q=X W_Q,\ K=X W_K, \ V=X W_V$,
	where the projections $W_Q$, $W_K$, and $W_V$ are all $d_k\times d$ parameter matrices. 
	The energy scores $E$ between any queries and keys are computed as \footnote{$Q K^{T}/\sqrt{d} $, the scaling factor $\sqrt{d}$ is omitted for simplicity. } 
	\begin{equation}
	E= Q K^\top, 
	\label{eqn:compatial}
	\end{equation}
	where $E$ is an $N \times N$ weight matrix, on which a softmax function is applied to obtain the weights of the values. 
	The output is computed as a weighted sum of the values as
	\begin{equation}
	Z=\operatorname{Attention}\left(Q, K, V\right)=\operatorname{Softmax}\left( E \right) V.
	\label{eqn:weignted_sum}
	\end{equation}

	\subsection{Self-attention network for image captioning} 
\begin{figure}[!t] 
		\centering
		\includegraphics[width=2.7in]{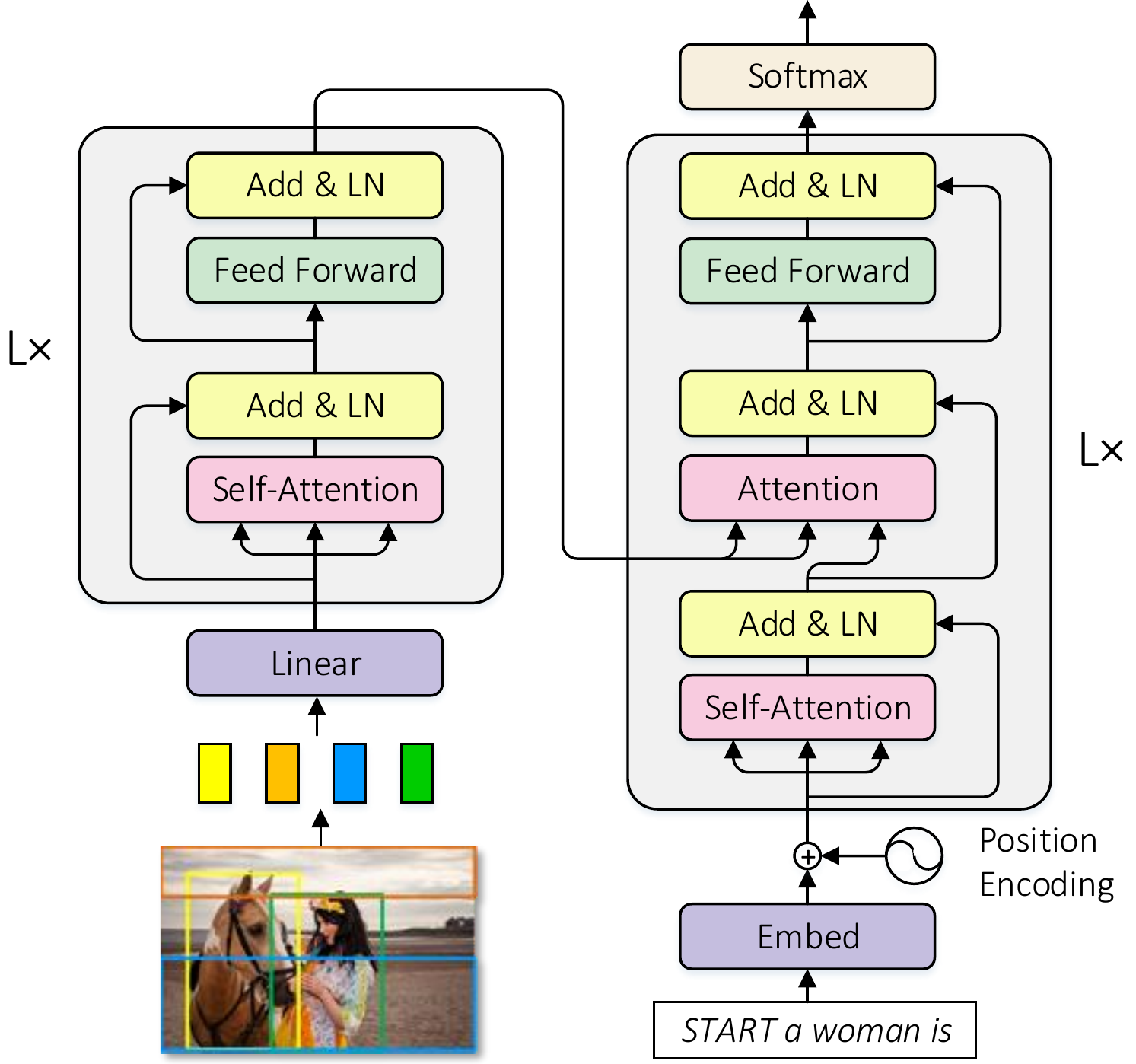}
		\caption{
			Architecture of the self-attention network (SAN) for image captioning.  		
		}
		\vspace{-0.3cm}
		\label{fig:tsfm}
	\end{figure}

	Figure~\ref{fig:tsfm} shows self-attention network (SAN), which is our baseline architecture for image captioning. 
	Similar to Transformer, the model consists of an image encoder and a caption decoder, both of which are composed of a stack of $L$ layers. 
	Each layer consists of one (for the encoder layer) or two (for the decoder layer) multi-head attention (MHA) sub-layers followed by a feed-forward network (FFN). 
	The MHA sub-layer contains $h$ parallel ``heads" with each head corresponding to an independent scaled dot-product attention function. 
	Besides, a residual connection and layer normalization are used between all the sub-layers. 
	
	The inputs to the encoder are the region-based visual features extracted from Faster-RCNN \cite{ren2015faster} object detector. 
	Each input element corresponds to an object in the image. 
	Before feeding the input vectors into the encoder, they are first passed through a dense layer followed by a ReLU layer to adapt their dimension to be consistent with the encoder. 
	The decoder takes the attended visual features and the embeddings of the previous words to predict the next word recursively. 
	Following Transformer, we add sinusoidal ``positional encodings" to 
	the inputs at the bottoms of the decoder. 
	Because the regions in the image don't have a natural order like sequences, no position information is added in the encoder side.

	\section{Approach}
	\subsection{Normalized SA (NSA)} 
	\label{sec:NSA}
	This section introduces a reparameterization of self-attention that takes advantage of normalization method for improved training. 
	
	We first review the formulation of Batch Normalization (BN). %
	Consider feeding an input mini-batch $x$ into a feed-forward layer $y=F(x, \Theta)$, where $F$ is an arbitrary transformation, and $\Theta$ is the parameter to be learned. 
	The internal covariate shift happens when the distribution of $x$ shifts during training. 
	To reduce internal covariate shift, BN normalizes each channel of $x$ using the mean and variance accumulated over the same channel in the whole mini-batch.

	We then take a closer look at the attention weight in Eqn.~\ref{eqn:weignted_sum}:  
	\begin{equation} 
	\begin{aligned} 
	S&=\operatorname{Softmax}(Q K^\top) \\ 
	 &= \operatorname{Softmax} ((XW_Q) \cdot (W_K^\top X^\top)) . 
	\end{aligned} 
	\label{eqn:att_score}
	\end{equation}
	It can be considered as an input instance $X\in \mathbb{R}^{N \times d_k}$ first goes through a $d_k \times d$ linear layer parameterized by $W_Q $ to obtain $Q=XW_Q\in \mathbb{R}^{N \times d}$, 
	which is then further fed into a $d\times N$ linear layer parameterized by $K^\top = W_K^\top X^\top$ followed by a Softmax activation to output $N$ probabilities over the keys. 
	Thus, we can re-formulate Eqn.~\ref{eqn:att_score} as a fully-connected layer $F$ followed by a Softmax activation: 
	\begin{equation} 
	\begin{aligned} 
	S &= \operatorname{Softmax}(F (Q,  \Theta)), \\
	Q&=XW^Q, \ \ \ \Theta = K^\top= W_K^\top X^\top.    
	\end{aligned} 
	\label{eqn:ff}
	\end{equation}
	
	Note that the parameter $\Theta$ is \textit{dynamically} calculated based on $X$. 
	From this perspective, SA can be susceptible to the internal covariate shift problem just as in a standard feed-forward network. 
	That is, when the distribution of input $Q$ shifts due to the change in network parameters during training,   
	the layer parameter $\Theta$ needs to continuously adapt to the new input distribution. 
	Consequently, SA may not be learned effectively.

	Therefore, to eliminate the internal covariate shift, it is advantageous for the distribution of $Q$ to remain fixed over time. 
	Then $\Theta$ does not have to readjust to compensate for the change in the distribution of $Q$.
	This can be accomplished by performing normalization on $Q$ by 
	\begin{equation} 
	\hat{Q}=\operatorname{Norm}(Q). 
	\end{equation}
	
	We now consider the implementation of $\operatorname{Norm}$. 
	BN is not directly suitable for $\operatorname{Norm}$ because instead of using a shared layer parameters for all examples in the dataset, 
	the layer parameter $\Theta=W_K^\top X^\top$ is dynamically computed with the \textit{instance-specific} $X$. 
	Therefore, it is more desirable to perform normalization, $\operatorname{Norm}$, for every single instance independently. 
	
	Let $x \in \mathbb{R}^{ B \times T \times C}$ and $x_{btc}$ denote the $btc-$th element of $x$, 
	where $b$ is the sample index, $c$ is the channel index, and $t$ is the index of the additional spatial dimension. 
	We implement $\operatorname{Norm}$ as normalizing each instance in the mini-batch independently using per-channel feature statistics: 
	\begin{equation} 
	\begin{aligned} 
	\hat{x}_{btc} &=\frac{x_{btc}-\mu_{bc}}{\sqrt{\sigma_{bc}^{2}+\epsilon}},  \\ %
	\mu_{b c} =\frac{1}{T} \sum_{t=1}^{T}x_{b t c },  & \ \ \ 
	\sigma_{b c}^{2} =\frac{1}{T} \sum_{t=1}^{T} \left(x_{b t c}-\mu_{b c} \right)^{2}. 
	\end{aligned}
	\label{eqn:in}
	\end{equation}
	
	The above normalization method is exactly the Instance Normalization (IN) in the 1D case. 
	Subtracting the mean from the queries could be considered as highlighting the differences among the queries and encourage them to query information from distinctive aspects. 

	We represent the normalization operation in Eqn.~\ref{eqn:in} as $\hat{x} = \operatorname{IN}(x)$. 
	Finally, we derive our normalized self-attention that reparameterizes the self-attention as 
	\begin{equation}
	\hat{Q}  = \operatorname{IN}(Q),\  \ \ \ 
	Z = \operatorname{Softmax}(\hat{Q} K^\top) V. 
	\label{eqn:norm}
	\end{equation}
	
	Similar to BN and IN, it is optional to further apply the channel-wise affine transformation  
	$\tilde{x}_{btc} =  \hat{x}_{btc} \gamma_c+\beta_c$ in $\operatorname{Norm}$, 
	where $\gamma, \beta \in \mathbb{R}^{C}$ are learnable scale and shift parameters. 
	But we empirically found it not necessary in our experiments. 
	It is also optional to normalize $K$ with $\hat{K}  = \operatorname{IN}(K)$.
	This is equivalent to normalizing the dynamic parameters $\Theta$, 
	which, however, may limit the capacity of SA.

	\paragraph{Relation to prior works.} 
	Our normalization method differs from Layer Normalization (LN) in that 
	LN normalizes along all channels of each individual element, 
	while our method normalizes along each channel of all input elements in an instance.  
	As for IN, it is typically used in 2D CNNs, \eg on style transfer task. 
	To our knowledge, IN has not been successfully used for language generation tasks, in particular for SAN.

	\subsection{Geometry-Aware SA (GSA)} 
	\label{sec:GSA}
	
	The inherent geometric structure among the input objects is beneficial for reasoning about the visual information, 
	which, however, is not modeled in the vanilla Transformer. 
	Therefore, we propose GSA that improves the SA module by taking into account the pairwise geometry relationships and the content information of objects.

	Denote the relative geometry features between two objects $i$ and $j$ as $\mathbf{f}^g_{ij}$, 
	which is a 4-dimensional vector of the relative position and size of the bounding boxes: 
	\begin{equation}\small
	\left(\log (\frac{ |x_i-x_j |}{w_i} ), \log (\frac{ |y_i-y_j|}{h_i} ), \log (\frac{w_i}{w_j} ), \log ( \frac{h_i}{h_j} ) \right)^{T},
	\end{equation}
	where $(x_i, y_i), w_i, h_i$ are the center coordinate, width, and height of box $i$, respectively. 
	
	We project $\mathbf{f}^g_{ij}$ to a high-dimensional representation $G_{ij}$ 
	with a fully-connected (FC) layer followed by a ReLU activation as 
	\begin{equation}
	G_{ij} = \operatorname{ReLU}\left(  \operatorname{FC} \left(  \mathbf{f}^g_{ij} \right) \right), 
	\end{equation}
	where $G \in \mathbb{R}^{N \times N \times d_g} $.

	We then modify the energy score in Eq.~\ref{eqn:compatial} to include the effect of $G$ as  
	\begin{equation}
	E = Q K^\top + \phi(Q^\prime, K^\prime, G), 
	\label{eqn:geometric}
	\end{equation}
	where $\phi$ is the geometric attention function, which outputs a score matrix of shape $N\times N$, and  
	$Q^\prime, K^\prime \in \mathbb{R}^{N \times  d_g}$ are geometric queries and keys that 
	are computed in the same way as $Q, K$, \ie by projecting the input $X$. 
	In the above equation, the first term is related to the queries and keys, namely \textit{content-based weight}. 
	The second term represents the \textit{geometric bias}, 
	which involves the geometry relations and the contents of $Q^\prime$ and $K^\prime$.

	We now discuss three choices of $\phi$, which can be either used individually or combined.
	
	\paragraph{Content-independent geometric bias.}
	The geometry relation $G_{ij}$ conveys useful information for understanding the relationships between two objects, \eg 
	object $i$ and $j$ have ``comparable sizes" and object $i$ is ``next to" object $j$. 
	Thus, we directly project $G_{ij}$ to a scalar score by 
	\begin{equation}
	\phi^1_{ij}  = \operatorname{ReLU}( w_g^\top G_{ij}), 
	\end{equation}
	where $w_g$ is the parameter to be learned. 
	The ReLU nonlinearity acts as a zero trimming operation so that 
	only the relations between objects with certain geometric relationships are considered. 
	
	The relation network \cite{hu2018relation} presented recently for object detection is a special case of the content-independent geometric bias. 
	Different from the above formulation, it fuses the content-independent geometric bias and the original attention weights by multiplication and use sinusoidal embedding of the geometry feature. 
	
	\paragraph{Query-dependent geometric bias.} 
	The above ``content-independent" variant assumes a \textit{static} geometric bias, \ie the same geometric bias is applied to all the query-key pairs in an SA layer. 
	However, the geometric biases are more often different, depending on what the associated query object is. 
	For example, for the queries, ``sea" and ``ball", their scale difference are often huge in the image, 
	and thus their sensitivities to the same change of a key's distance/position vary widely. 
	Therefore, the geometric biases of the two queries should be adapted to match their content. 
	To this end, we decide to \textit{dynamically} compute the geometric bias for different queries: 
	\begin{equation}
	\phi^2_{ij}  = {Q^\prime}_{i}^\top G_{ij}. 
	\end{equation}
	Here we use dot-product to match ${Q^\prime}_{i}$ with $G_{ij}$ 
	since it is more computation and memory efficient than using the Concatenation-FC operation. 
	
	\paragraph{Key-dependent geometric bias.} 
	Similar to the query-dependent variant, geometric bias can also be associated with the content of the keys, computed as
	\begin{equation}
	\phi^3_{ij}  = {K^\prime}_{j}^\top G_{ij}. 
	\end{equation}

	\subsection{Applying NSA and GSA modules to SAN} 
	We first combine both NSA and GSA by replacing $Q$ in Eqn.~\ref{eqn:geometric} with the normalized one, $\hat{Q}$. 
	We then use this module to replace the vanilla SA modules in the encoder of SAN, which 
	results in our full model, namely Normalized and Geometry-aware Self-Attention Network (\textbf{NG-SAN}). 
	NSA is not applied in the decoder of SAN because the decoder is autoregressive and 
	has variable-length inputs. 
	This is undesirable for IN because the mean and variance statistics are meaningless when the sequence length is 1.

	\section{Experiments on Image Captioning}
	\subsection{Experimental setup} 
	\paragraph{MS-COCO dataset \cite{lin2014microsoft}. } 
	It is the most popular benchmark for image captioning. 
	We use the `Karpathy' splits that have been used extensively for 
	reporting results in prior works. This split contains 113,287
	training images with 5 captions each, and 5k images for validation and test splits, respectively. 
	We follow standard practice \cite{vinyals2015show} to pre-process the text, 
	resulting in a final vocabulary of 9,487 words. 
	We use the region-based image features provided by Bottom-Up \cite{anderson2017bottom} for training.

	\begin{figure}[!t] 
		\centering
		\includegraphics[width=2.7in]{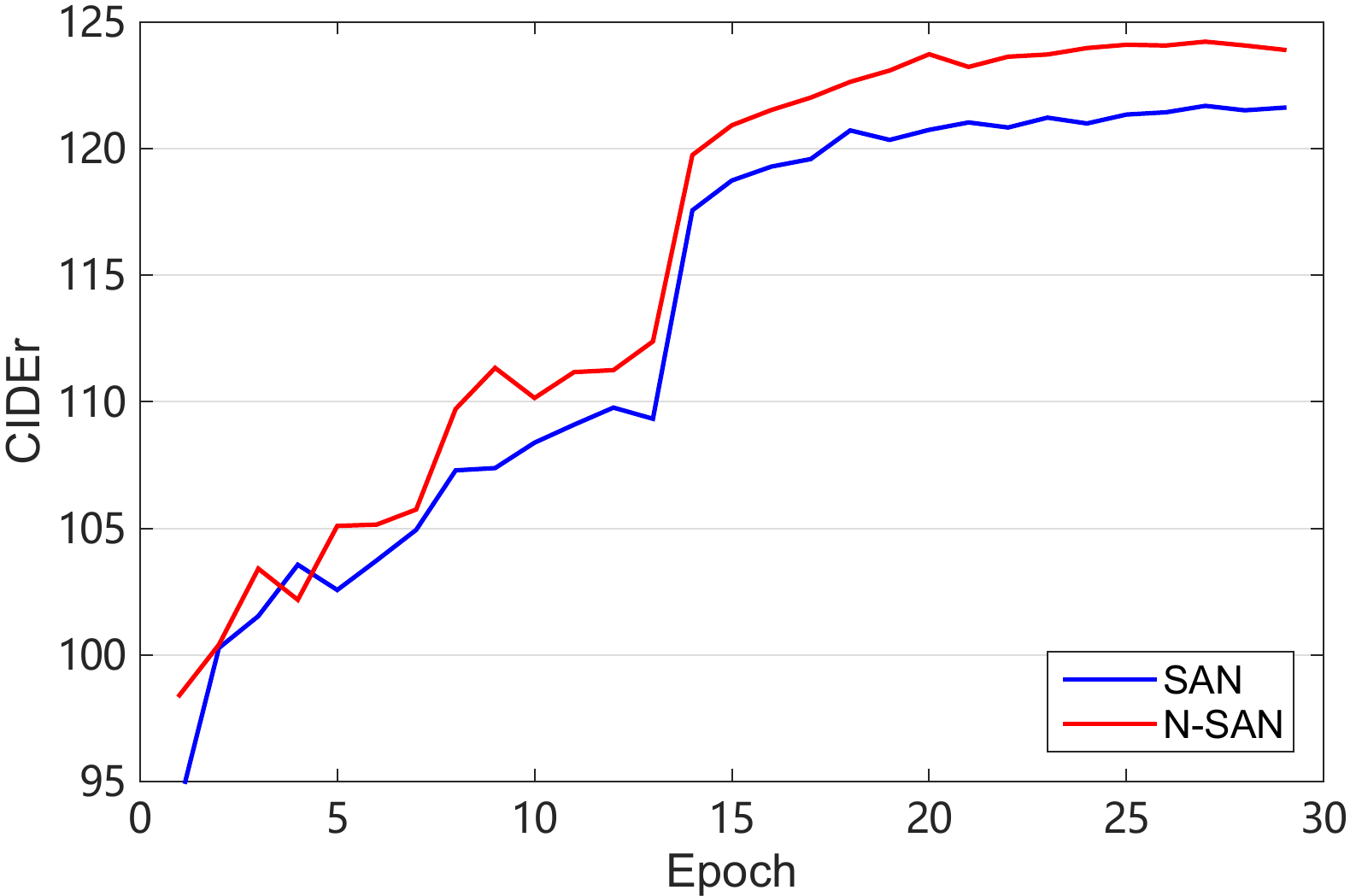}
		\caption{
			Changes of CIDEr scores during training.  		
		}
	\vspace{-0.3cm}
		\label{fig:curve}
	\end{figure}
	
	\paragraph{Evaluation metrics. }
	We use the standard automatic evaluation metrics to evaluate the quality of image captions, including BLEU-1/2/3/4 \cite{Papineni2002BLEU}, METEOR \cite{Denkowski2014Meteor}, ROUGE-L \cite{lin2004rouge}, CIDEr \cite{Vedantam2015CIDEr}, and SPICE \cite{Anderson2016SPICE}, which are denoted as B@1/2/3/4, M, R, C and S, respectively. 
	
	\vspace{-0.3cm}
	\paragraph{Implementation details.}
	We follow Transformer-\textit{Base} model \cite{vaswani2017attention} and \cite{yu2019multimodal} to set the model hyper-parameters and train the model. 
	Specifically, the dimensionality of input image features is 2048. 
	The latent dimension in the MHA module is 512, and the number of heads is 8. 
	Inner dimension in the FFN module is 2,048. 
	We apply \textit{dropout} with a probability of 0.1. 
	We use the same number of layers $L$ for the encoder and decoder. %
	For training, we use the Adam optimizer \cite{kingma2014adam} 
	We use a step decay schedule with warm-up for varying the learning rate. 
	The base learning rate is set to 
	$min( t\times 10^{-4}; 3\times 10^{-4})$, where $t$ is the current epoch number that starts at 1. 
	After 6 epochs, the learning rate is decayed by 1/2 every 3 epochs. 
	All models are first trained for 15 epochs with the cross-entropy loss 
	and then further optimized with CIDEr reward \cite{Rennie2016Self} for additional 15 epochs. 
	If not specifically mentioned, by default we set $L=4$, only normalize the query and do not apply $\gamma, \beta$ in NSA, 
	and use the query-dependent variant ($\phi^1$) of GSA. 
	Beam search with a beam width of 3 is used during testing stage.

	\subsection{Analysis on NSA} 
	In this section, we examine the effectiveness of NSA module. 
	We replace the SA modules in the encoder of SAN with NSA, resulting in a model named 
	Normalized Self-Attention Network (\textbf{N-SAN}). 
	
	\paragraph{Number of attention layers.} 
	In Table~\ref{tab:norm_layers} we compare the performance of N-SAN and SAN  
	under the same number of SA layers $L\in \{1,2,4,6\}$.  
	We can see that the model size grows linearly as $L$ increases. 
	Regarding the performance, we have two observations as follows. 
	1) As $L$ increases, the performance of both SAN and N-SAN gradually improves and reaches the optimal value when $L=6$. 
	However, the performance gain of increasing $L$ from 4 to 6 is not very significant. 
	Therefore, we use $L=4$ for later experiments as a compromise between the model's performance and complexity.  
	2) N-SAN consistently outperforms SAN on all metrics under different $L$. 
	In Figure~\ref{fig:curve}, we further plot the CIDEr scores of the one-layer SAN and N-SAN models during training, evaluated on the validation split at each epoch. 
	As we can see, the curve of N-SAN is above that of SAN for most of the time.

		\begin{table}[tbp]
		\centering
		\caption{Comparisons between N-SAN and SAN using different numbers of self-attention layers $L$.}
		\resizebox{0.47\textwidth}{!}{
			\begin{tabular}{clrrrrrr}
				\toprule
				\#Layers & Model & \multicolumn{1}{c}{\#params\newline{}} & \multicolumn{1}{c}{B@4\newline{}} & \multicolumn{1}{c}{M\newline{}} & \multicolumn{1}{c}{R} & \multicolumn{1}{c}{C} & \multicolumn{1}{c}{S\newline{}} \\
				\midrule
				\multirow{2}[2]{*}{$1$} 
				& SAN & 18.1M  & 36.8 & 28.0 & 57.6 & 123.4 & 21.8 \\
				& N-SAN & 18.1M & \bf 38.2 & \bf 28.6 & \bf  58.2 & \bf 127.2 & \bf 22.2 \\
				\midrule
				\multirow{2}[2]{*}{$2$} 
				& SAN & 25.5M & 38.2 & 28.5 & 58.3 & 127.1 & 22.3 \\
				& N-SAN & 25.5M & \bf 38.9 & \bf 28.9 & \bf 58.6 & \bf  129.7 & \bf 22.6 \\
				\midrule
				\multirow{2}[2]{*}{$4$} 
				& SAN & 40.2M & 38.4 & 28.6 & 58.4 & 128.6 & 22.6 \\
				& N-SAN & 40.2M  & \bf 39.3 &\bf  29.1 &\bf  58.9 & \bf 130.8 &\bf  23.0 \\
				\midrule
				\multirow{2}[2]{*}{$6$} 
				& SAN & 54.9M& 38.6 & 28.6 & 58.5 & 128.8 & 22.5 \\
				& N-SAN & 54.9M& \bf 39.3 & \bf 29.2 & \bf 59.1 & \bf 131.1 & \bf 23.0 \\
				\bottomrule
			\end{tabular}%
		}
		\label{tab:norm_layers}%
	\end{table}%
	
	\begin{table}[tbp]
	\centering
	\caption{Comparison of using various normalization methods in NSA. }
	\resizebox{0.39\textwidth}{!}{
		\begin{tabular}{lrrrrrr}
			\toprule
			Approach\newline{}  & \multicolumn{1}{c}{B@4\newline{}} & \multicolumn{1}{c}{M\newline{}} & \multicolumn{1}{c}{R} & \multicolumn{1}{c}{C} & \multicolumn{1}{c}{S\newline{}} \\
			\midrule
			SAN & 38.4 & 28.6 & 58.4 & 128.6 & 22.6\\
			\midrule
			LN  &  38.5 & 28.6 & 58.3 & 128.2 & 22.5 \\
			BN  &  38.8 & 28.9 & 58.7 & 129.4 & 22.8 \\
			IN   &  \bf 39.4 & \bf 29.2 & \bf 59.0 & 130.7 & \bf 23.0\\
			IN  w/o $\gamma, \beta$ & 39.3 &29.1 &58.9 & \bf130.8 & \bf 23.0 \\
			\bottomrule
		\end{tabular}%
	}
	\vspace{-0.3cm}
	\label{tab:norm_type}%
	\end{table}%

	\vspace{-0.2cm}
	\paragraph{Different normalization methods.} 
	Since we introduced IN into the NSA module for normalization,  
	an intuitive question to ask is whether we can replace IN with other normalization methods. 
	In Table~\ref{tab:norm_type} we show the results of using different normalization methods including BN, LN, IN and IN without using the affine transformations ($\gamma$ and $\beta$).  
	We have the following observations. 
	1) Using LN slightly decreases the performance. 
	We conjecture that is because LN normalizes activations of all channels with the same normalization terms ($\mu$ and $\sigma$), 
	thus limiting the expression capacity of each channel when calculating attention weights. 
	2) IN and \textit{IN  w/o $\gamma, \beta$} significantly outperform SAN and all the other normalization methods. Meanwhile, the extra affine transformations ($\gamma$ and $\beta$) are not necessary. 
	3) Applying BN outperforms SAN but is inferior to adopting IN. 
	BN has a similar effect as IN to reduce the internal covariate shift by fixing the distribution of the queries. 
	However, as is described in Sec.~\ref{sec:NSA}, 
	since the layer parameter $\Theta$ in Eqn.~\ref{eqn:ff} depends on instance-specific input, it is more desirable to perform input normalization also on each instance instead of on the whole mini-batch.

	\paragraph{What if we normalize the keys in addition to the queries?} 
	In Table~\ref{tab:norm_which}, we compare the variants of Eqn.~\ref{eqn:norm}, including 
	normalizing Q alone, K alone, and both Q and K. 
	We have the following observations. 
	1) Normalizing either of Q and K could increase the performance. 
	2) The performances of normalizing both Q and K and normalizing Q alone are very similar, 
	and are both significantly higher than that of SAN. 
	3) Normalizing K alone is inferior to normalizing Q alone. 
	The reason is that normalizing $K$ is equivalent to normalizing $\Theta$ in Eqn.~\ref{eqn:ff},  
	which may limit the model capacity of SA.

	\begin{table}[tp]
		\centering
		\caption{Comparison of normalizing query and key in N-SAN.  }
		\resizebox{0.38\textwidth}{!}{
			\begin{tabular}{cc|rrrrrr}
				\toprule
				Query & Key  & \multicolumn{1}{c}{B@4\newline{}} & \multicolumn{1}{c}{M\newline{}} & \multicolumn{1}{c}{R} & \multicolumn{1}{c}{C} & \multicolumn{1}{c}{S\newline{}} \\
				\midrule
				\xmark &\xmark & 38.4 & 28.6 & 58.4 & 128.6 & 22.6 \\
				\cmark &\xmark   &  39.3 & \bf 29.1 &\bf 58.9 & \bf 130.8 &23.0 \\
				\xmark&\cmark  &   39.2 & 29.0 & 58.8 & 130.1 & 22.8 \\
				\cmark& \cmark  & \bf 39.4 &\bf  29.1 & 58.8 &  130.7 & \bf 23.1 \\
				\bottomrule
			\end{tabular}%
		}
		\label{tab:norm_which}%
	\end{table}%
	
	\begin{table}[tbp]
	\centering
	\caption{Comparison of various variants of GSA. }
	\resizebox{0.49\textwidth}{!}{
		\begin{tabular}{lrrrrrr}
			\toprule
			Approach\newline{}  & \multicolumn{1}{c}{\#params\newline{}} & \multicolumn{1}{c}{B@4\newline{}} & \multicolumn{1}{c}{M\newline{}} & \multicolumn{1}{c}{R} & \multicolumn{1}{c}{C} & \multicolumn{1}{c}{S\newline{}} \\
			\midrule
			SAN & 40.2M & 38.4 & 28.6 & 58.4 & 128.6 & 22.6\\
			\midrule
			absolute & 40.2M & 38.3 & 28.5 & 58.4 & 128.4 & 22.6\\
			content-independent & 40.2M  & 39.2& 29.1& 58.9 & 131.0 & 22.9  \\
			key-dependent  &  41.5M  & 38.9 & 29.0 & 58.8 & 129.5 & 22.8\\
			query-dependent  &  41.5M & \bf 39.3 & \bf 29.2 & \bf 59.0 & \bf 131.4 & \bf 23.0\\
			\bottomrule
		\end{tabular}%
	}
	\vspace{-0.3cm}
	\label{tab:gsa_type}%
\end{table}%

	\begin{table*}[tbp]
		\centering
		\caption{Leaderboard of the published state-of-the-art, \textit{single-model} methods on the online MS-COCO test server, where c5 and c40 denote using 5 and 40 references for testing, respectively. CIDEr (C40) is the default sorting metric on the leaderboard. 
		}
		\resizebox{\textwidth}{!}{
			\begin{tabular}{@{\extracolsep{3pt}}@{\kern\tabcolsep}lrrrrrrrrrrrrrr}
				\toprule
				\multicolumn{1}{c}{\multirow{2}[4]{*}{Model}}  & \multicolumn{2}{c}{BLEU-1} & \multicolumn{2}{c}{BLEU-2} & \multicolumn{2}{c}{BLEU-3} & \multicolumn{2}{c}{BLEU-4} & \multicolumn{2}{c}{METEOR} & \multicolumn{2}{c}{ROUGE-L} & \multicolumn{2}{c}{CIDEr-D} \\
				\cmidrule{2-3}  \cmidrule{4-5} \cmidrule{6-7} \cmidrule{8-9} \cmidrule{10-11} \cmidrule{12-13} \cmidrule{14-15} %
				& \multicolumn{1}{c}{c5} & \multicolumn{1}{c}{c40} & \multicolumn{1}{c}{c5} & \multicolumn{1}{c}{c40} & \multicolumn{1}{c}{c5} & \multicolumn{1}{c}{c40} & \multicolumn{1}{c}{c5} & \multicolumn{1}{c}{c40} & \multicolumn{1}{c}{c5} & \multicolumn{1}{c}{c40} & \multicolumn{1}{c}{c5} & \multicolumn{1}{c}{c40} & \multicolumn{1}{c}{c5} & \multicolumn{1}{c}{c40} \\
				\midrule
				Up-Down \cite{anderson2017bottom} & 80.2  &\bf 95.2  & 64.1  & 88.8  & 49.1  & 79.4  & 36.9  & 68.5  & 27.6  & 36.7  & 57.1  & 72.4  & 117.9  & 120.5  \\
				CAVP \cite{liu2018context}  & 80.1  & 94.9  & 64.7  & 88.8  & 50.0  & 79.7  & 37.9 & 69.0  & 28.1  & 37.0  & 58.2  & 73.1  & 121.6  & 123.8  \\
				SGAE \cite{yang2019auto} & 80.6  & 95.0  & 65.0  & 88.9  & 50.1  & 79.6  & 37.8  & 68.7  & 28.1  & 37.0  & 58.2  & 73.1  & 122.7  & 125.5  \\
				VSUA \cite{guo2019vsua} & 79.9  & 94.7  & 64.3  & 88.6  & 49.5  & 79.3  & 37.4  & 68.3  & 28.2  & 37.1  & 57.9  & 72.8  & 123.1  & 125.5  \\
				\midrule
				NG-SAN (Ours) & \bf 80.8 & 95.0 & \bf65.4 & \bf89.3 & \bf50.8 & \bf80.6 & \bf38.8 & \bf70.2 & \bf29.0 & \bf38.4 & \bf58.7 & \bf74.0 & \bf126.3 & \bf 128.6 \\
				\bottomrule
			\end{tabular}%
		}
	\vspace{-0.2cm}
		\label{tab:server}%
	\end{table*}%
	
	\subsection{Analysis on GSA} 
	In this section, we examine the effectiveness of GSA module. 
	Similar to N-SAN, we replace the SA modules in the encoder of SAN with GSA to obtain a model named 
	Geometry-aware Self-Attention Network (\textbf{G-SAN}). 
	
	\vspace{-0.2cm}
	\paragraph{Variants of GSA.} 
	In Table~\ref{tab:gsa_type} we compare various variants of GSA module introduced in Sec.~\ref{sec:GSA}. 
	``+absolute" denotes adding absolute geometry information 
	of each individual object to their input representations at the bottoms of the encoder.  
	It is obtained by embedding the geometry features, \ie 
	the center coordinates and the width/height of the box, normalized by the width/height of the image,  
	to a sinusoidal representation using the same method as the ``positional encodings" in \cite{vaswani2017attention}. 
	We have the following findings. 
	1) Adding the absolute geometry information (``absolute") is not beneficial to the performance.  
	That is probably because it is too complex for SA to infer the 2D layout of objects from their absolute geometry information. 
	2) All the proposed variants of GSA can improve the performance of SAN, showing the advantages of using relative geometry information. 
	3) ``query-dependent" brings the best performance and outperforms the content-independent variant, 
	proving that incorporating the content information of the associated query can help infer a better geometric bias. 
	4) ``key-dependent"  is inferior to ``query-dependent". 
	That is because when using key-dependent geometric bias, the scores $\phi^3_{ij}  = {K^\prime_{j}}^\top G_{ij}$ condition on \textit{different} keys $K^\prime_{j}$, thus the differences in $G_{ij}$ may be overwhelmed by the differences in $K^\prime_{j}$ when performing softmax on the keys' dimension. 
	In comparision, when using query-dependent geometric bias, the effect of $G_{ij}$ could be highlighted since the scores condition on a \textit{common} query ${Q^\prime_{i}}$ when performing softmax. 
	We did not observe further improvement when combing these variants into $\phi$ in Eq. \ref{eqn:geometric}.

	\subsection{Analysis on the full model (NG-SAN)}
	We now validate the effectiveness of NG-SAN that takes advantage of both NSA and GSA. 
	
	\vspace{-0.2cm}
	\paragraph{Comparisons with state-of-the-arts.}
	We compare NG-SAN with the state-of-the-art methods, including Up-Down \cite{anderson2017bottom}, 
	CAVP \cite{liu2018context}, SGAE \cite{yang2019auto}, VSUA \cite{guo2019vsua}, ORT \cite{herdade2019image}, AoANet \cite{huang2019attention}, and MT \cite{yu2019multimodal}. 
	All the methods except ORT, AoANet, and MT are based on single- or multi-layer Long Short-Term Memory (LSTM) networks. 
	MT adopts a Transformer-Base architecture, using 6 SA layers for both the encoder and the decoder, 
	and inserts an additional LSTM layer in the decoder. 
	ORT also adopts the Transformer-Base architecture and follows \cite{hu2018relation} to model the spatial relationship between inputs. 
	AoANet uses SAN as the encoder and LSTM as the decoder.

	Table~\ref{tab:test} compares the results of each method. 
	We can see that both G-SAN and N-SAN outperform the SAN baseline across all metrics. 
	Moreover, NG-SAN further outperforms G-SAN and N-SAN, demonstrating that GSA and NSA are compatible with each other. 
	NG-SAN significantly outperforms all the other methods, including both LSTM-based and SA-based ones, over all metrics. 
	Particularly, we improve the best CIDEr score from 130.9 to 132.1. 
		Table~\ref{tab:server} further reports the performance of the top-performing \textit{single-model} solutions on the official test server. %
	Compared with the published methods, 
	our single model significantly outperforms all the 
	other methods in terms of all evaluation metrics except BLEU-1. 
	In particular, we establish a new state-of-the-art score of 128.6 on CIDEr (C40). 
			
	\begin{table}[tbp]
		\centering
		\caption{Comparisons with state-of-the-art single-model approaches on MS-COCO Karpathy test split.  }
		\resizebox{0.48\textwidth}{!}{
			\begin{tabular}{lcccccc}
				\toprule
				\multicolumn{1}{c}{Model} & \multicolumn{1}{c}{\#params}   & \multicolumn{1}{c}{B@4} & \multicolumn{1}{c}{M} & \multicolumn{1}{c}{R} & \multicolumn{1}{c}{C} & \multicolumn{1}{c}{S} \\
				\midrule
				Up-Down \cite{anderson2017bottom} & --& 36.3 &27.7 &56.9& 120.1& 21.4 \\
				CAVP \cite{liu2018context}  & -- & 38.6 & 28.3&  58.5&  126.3&  21.6 \\
				SGAE \cite{yang2019auto}  & --   & \multicolumn{1}{c}{39.0} & \multicolumn{1}{c}{28.4} & \multicolumn{1}{c}{58.9} & \multicolumn{1}{c}{129.1} & \multicolumn{1}{c}{22.2} \\
				VSUA \cite{guo2019vsua} & --& 38.4  &  28.5  &  58.4  &  128.6  & 22.0 \\
				ORT \cite{herdade2019image} & -- & 38.6 & 28.7 & 58.4 & 128.3 & 22.6 \\
				AoANet \cite{huang2019attention} & -- & 38.9& 29.2 & 58.8 & 129.8 & 22.4 \\
				MT \cite{yu2019multimodal}  & 57.0M & \multicolumn{1}{c}{39.8} & \multicolumn{1}{c}{29.1} & \multicolumn{1}{c}{59.1} & \multicolumn{1}{c}{130.9} &  --\\
				\midrule
				SAN  &  40.2M   & 38.4 & 28.6 & 58.4 & 128.6 & 22.6    \\
				N-SAN &   40.2M  &  39.3 & 29.1 & 58.9 & 130.8 & 23.0 \\
				G-SAN &   41.5M  &  39.3 & 29.2&  59.0 & 131.4 & 23.0 \\
				NG-SAN &  41.5M  &   \bf39.9  &   \bf  29.3  &   \bf  59.2  &   \bf 132.1   & \bf23.3 \\
				\bottomrule
			\end{tabular}%
		}
		\label{tab:test}%
	\end{table}%

	\vspace{-0.2cm}
	\paragraph{Complexity.} 
	As can be seen in the ``\#params" column in Table~\ref{tab:test}, NG-SAN requires very few (about 2k) additional parameters compared with SAN. 
	For NSA, it does not require any parameters, and the computation overhead of the additional normalization process is almost ignorable.  
	While GSA indeed requires some additional parameters, the amount is ignorable. 
	GSA can be efficiently implemented by matrix multiplication and the einstein summation (einsum) operations provided by mainstream deep learning frameworks.

	\section{Extension: Experiments on Other Tasks}
	We further investigate the effectiveness and generality of our methods on Video Captioning (VC) \cite{venugopalan2015sequence}, Machine Translation (MT) \cite{bahdanau2014neural}, and Visual Question Answering (VQA) \cite{antol2015vqa} tasks. 
	Since VC and MT are both sequence-to-sequence problems, 
	we directly use Transformer as the baseline models,   
	and we replace the SA modules in their encoder with the proposed NSA module to construct our methods. 
	As for VQA, we use MCAN \cite{Yu_2019_CVPR} as the baseline model, which uses a SAN-based network to simultaneously encode image and question information. 
	To build our method for VQA, we replace all the SA modules in MCAN with our GSA modules.  
	
	\subsection{Video Captioning} 
	We use a recently released large-scale video captioning dataset, VATEX \cite{wang2019vatex}.
	It contains over 41,250 videos and 412,500 English captions. 
	For a fair comparison with VATEX, we directly use the pre-extracted video features provided by the paper. 
	Specifically, each video is sampled at 25fps and 1,000-dimensional features are extracted from these sampled frames using a pretrained I3D \cite{carreira2017quo} model. 
	Because the dataset is relatively small, we found using one layer in both the encoder and decoder is satisfactory.   
	We use a training configuration the same as that of our image captioning model. 
	
	In Table~\ref{tab:video}, we compare our method with the Transformer baseline and the VATEX model. 
	We see that the performance of Transformer strongly exceeds that of VATEX, which adopts an LSTM-based architecture.  
	Our Transformer+NSA method consistently improves over Transformer on all metrics. 
	Particularly, our method improves the CIDEr score by 3.7 points when compared to Transformer,  
	and significantly improves the CIDEr score by 11.4 points when compared to VATEX baseline. 
	
	\begin{table}[tbp]
		\centering
		\caption{Video captioning results on VATEX dataset. }
		\resizebox{0.38\textwidth}{!}{
			\begin{tabular}{lrrrrr}
				\toprule
				Model & \multicolumn{1}{c}{B@4\newline{}} & \multicolumn{1}{c}{M\newline{}} & \multicolumn{1}{c}{R} & \multicolumn{1}{c}{C}  \\
				\midrule
				VATEX \cite{wang2019vatex}&  28.2 & 21.7 & 46.9 & 45.7 \\
				\midrule
				Transformer (Ours) & 30.6 & 22.3 & 48.4 & 53.4 \\
				\ \ +NSA & \bf 31.0 & \bf22.7 &\bf 49.0 & \bf57.1 \\
				\bottomrule
			\end{tabular}%
		}
		\label{tab:video}%
	\end{table}%

	\begin{table}[tbp]
		\centering
		\caption{Machine translation results on newstest2014 for WMT 2014 En-De dataset. }
		\vspace{0.05cm}
		\resizebox{0.28\textwidth}{!}{ 
			\begin{tabular}{lr}
				\toprule
				Model & BLEU  \\
				\midrule
				Transformer-Base \cite{vaswani2017attention} & 27.30 \\
				Transformer-Big \cite{vaswani2017attention} & 28.40 \\
				\midrule
				Transformer-Base (Ours) & 27.56 \\
				\ \ +NSA &  \bf27.92 \\
				\bottomrule 
			\end{tabular}%
		}
	\vspace{-0.2cm}
		\label{tab:mt}%
	\end{table}%

	\subsection{Machine Translation} 
	We also evaluate NSA on MT task, for which the Transformer was originally proposed. 
	We trained on the widely-used WMT 2014 English to German (En-–De) dataset, which consists of about 4.56 million sentence pairs. 
	The models were validated on newstest-2013 and tested on newstest-2014 with BLEU. 
	We use the well-known Transformer-Base \cite{vaswani2017attention} variant of Transformer as the baseline model, 
	which has 6 layers in both the encoder and decoder. 
	Specifically, we follow the implementation of the \texttt{fairseq-py} \cite{ott2019fairseq} toolkit.

	As shown in Table~\ref{tab:mt}, 
	Compared to Transformer-Base model, NSA increases the BLEU score by 0.36 points without adding any parameters.

	\subsection{Visual Question Answering}
	We conduct experiments on the most commonly used VQA benchmark, VQA-v2 \cite{antol2015vqa}. 
	It contains human-annotated question-answer pairs relating to the images from the MS-COCO dataset, with 3 questions per image and 10 answers per question.
	We strictly follow MCAN \cite{Yu_2019_CVPR} to implement our models. 
	Specifically, images are represented with region features extracted from Faster R-CNN object detector 
	and the input questions are transformed with GloVe word embeddings and an LSTM network. 
	
	Table~\ref{tab:vqa} shows the overall accuracies of our methods and the current state-of-the-art models on the online test-dev and test-std splits. 
	GSA boosts the test-std accuracy of MCAN from 70.83 to 71.28. 
	\begin{table}[tbp]
		\centering
		\caption{Visual question answering accuracies on the  VQA-v2 dataset to compare with the state-of-the-art single-model methods. }
		\resizebox{0.28\textwidth}{!}{
			\begin{tabular}{lrrrrr}
				\toprule
				Model & \multicolumn{1}{c}{test-dev} & \multicolumn{1}{c}{test-std}  \\
				\midrule
				MLIN \cite{gao2019multi} & 70.18 & 70.28 \\
				DFAF \cite{Gao_2019_CVPR} & 70.22 & 70.34 \\
				MCAN \cite{Yu_2019_CVPR}&  70.63 & 70.90 \\
				\midrule
				MCAN (Ours) & 70.54 & 70.83 \\
				\ \ +GSA & \bf70.76 & \bf71.28 \\
				\bottomrule
			\end{tabular}%
		}
	\vspace{-0.3cm}
		\label{tab:vqa}%
	\end{table}%

	\section{Conclusion}
	We proposed two improvements to the self-attention (SA) mechanism, \ie 
	a Normalized Self-Attention (NSA) to reduce the internal covariate shift problem inside SA, and
	a class of Geometry-aware Self-Attention (GSA) that explicitly and dynamically computes the geometric bias between objects to benefit image understanding.
	We have conducted extensive experiments on MS-COCO image captioning dataset to validate the effectiveness of NSA, GSA, and their combination.
	We further show the significance and generality of our methods on video captioning, machine translation, and visual question answering tasks. 
	On all tasks, simply replacing the vanilla SA module with our proposed methods provides solid improvements over strong baselines.

	\vspace{-0.5cm}
	\paragraph{Acknowledgments} 
	This work was supported by National Natural Science Foundation of China (No.61922086 and No.61872366) and Beijing Natural Science Foundation (No.4192059).
	
	{\small
		\bibliographystyle{ieee_fullname}
		\bibliography{references}
	}

	\onecolumn

	\appendix
	\section{Appendix}

	\subsection{Visualization of Geometric Weights}
	To gain a better insight about the effect of the relative geometry information on attention weights, 
	we visualize the geometric weights in GSA. 
	Specifically, we use the content-dependent version ($\phi^1$) of GSA, and use a trained one-layer G-SAN model. 
	We visualize how the geometric weight $\phi^1_{ij}$ between object $i$ and $j$ changes as the relative geometry feature $\mathbf{f}^g_{ij}$ between them changes. 
	
	Review that the relative geometry feature $\mathbf{f}^g_{ij}$ is a 4-dimensional vector: 
	\begin{equation}\small
	\mathbf{f}^g_{ij} = \left(\log (\frac{ |x_i-x_j |}{w_i} ), \log (\frac{ |y_i-y_j|}{h_i} ), \log (\frac{w_i}{w_j} ), \log ( \frac{h_i}{h_j} ) \right)^{T},
	\end{equation}
	where $(x_i, y_i), w_i, h_i$ are the center coordinate, width, and height of box $i$, respectively. 
	We simplify the above equation as 
	\begin{equation}
	\mathbf{f}^g_{ij} = \left(\log ( \Delta x ), \log (\Delta y), \log (\Delta w), \log (\Delta h) \right)^{T}. 
	\end{equation}
	
	We then keep one of $(\Delta x, \Delta y)$ and $(\Delta w, \Delta h)$ fixed, change the other one, and see plot the values of $\phi^1_{ij}$.

	\begin{figure*}[b] 
		\centering
		\includegraphics[width=0.9\textwidth]{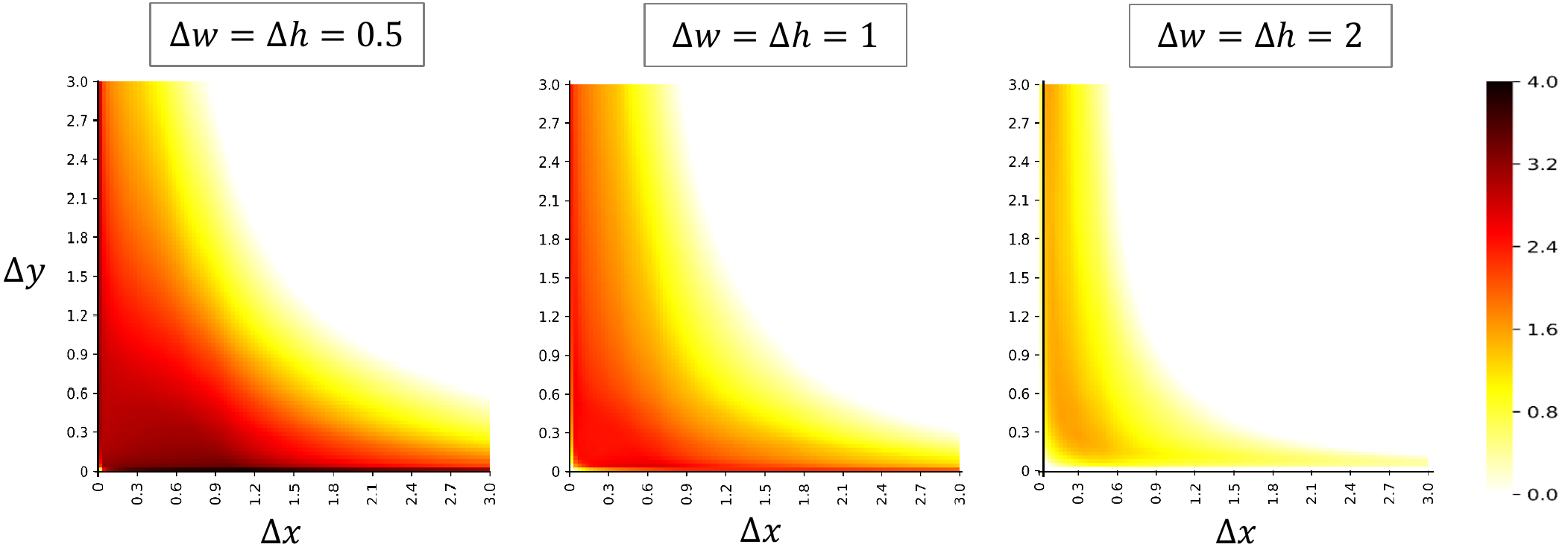}
		\caption{
			Visualization of the geometric weights as a function of the relative position. 
			Each plot shows the values of $\phi^1_{ij}$ as $\Delta x$ and $\Delta y$ changing in range of $[0,3]$, while $\Delta w=\Delta h \in \{0.5, 1, 2\}$ are kept fixed. 
		}
		\label{fig:xy}
	\end{figure*}
	\begin{figure*}[b] 
		\centering
		\includegraphics[width=0.9\textwidth]{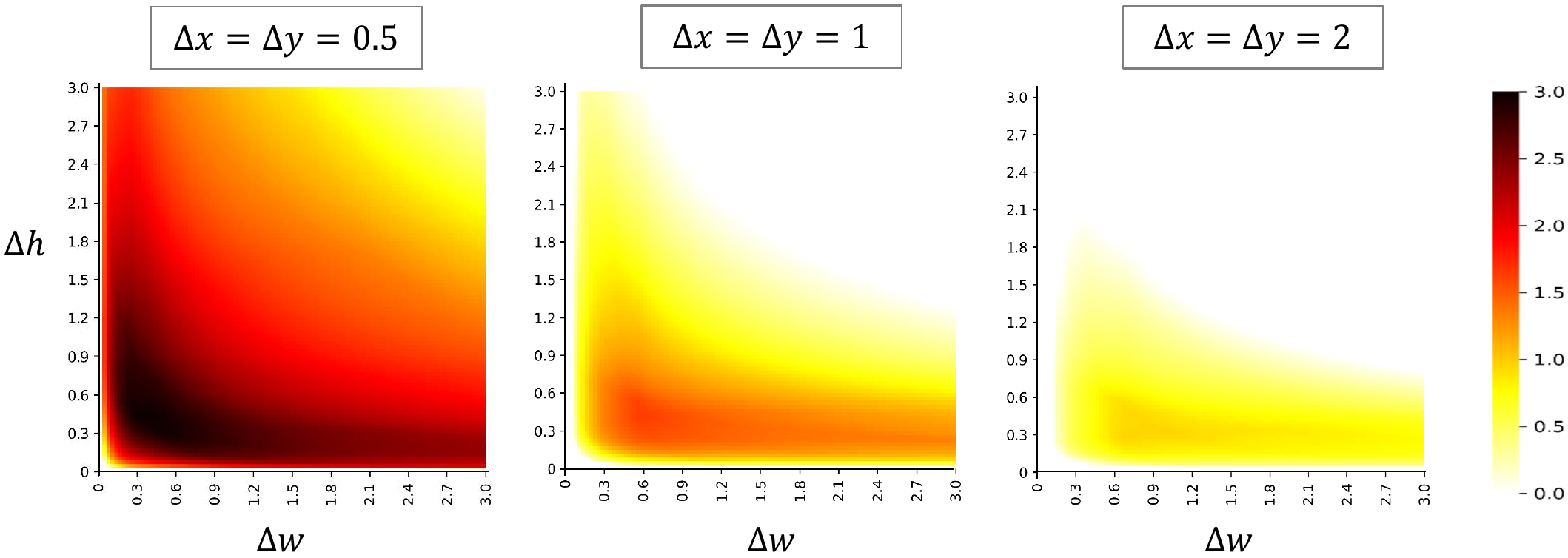}
		\caption{
			Visualization of the geometric weights as a function of the relative width and height. 
			Each plot shows the values of $\phi^1_{ij}$ as $\Delta w$ and $\Delta h$ changing in range of $[0,3]$, while $\Delta x=\Delta y \in \{0.5, 1, 2\}$ are kept fixed. 
		}
		\label{fig:wh}
	\end{figure*}
	
	Figure~\ref{fig:xy} shows the cases when $\Delta w=\Delta h \in \{0.5, 1, 2\}$ are fixed, and $\Delta x$ and  $\Delta y$ change in range of $[0,3]$. 
	We can observer that, basically, the geometric weight gets smaller when the relative distance between the two objects increases. 
	Exceptions are found near $(\Delta x, \Delta y)=(0,0)$, where the weights are relatively smaller than neighboring point. 
	That is probably because when two boxes $i$ and $j$ have similar sizes, \eg $\Delta w=\Delta h=1$, and their center coordinates almost coincide, 
	then they likely refer to the same object. 
	Therefore, the weight of box $j$ should be reduced to avoid repeating the object.

	Figure~\ref{fig:wh} shows the cases when $\Delta x=\Delta y \in \{0.5, 1, 2\}$ are fixed, and $\Delta w$ and  $\Delta h$ change in range of $[0,3]$. 
	We have the following observations. 
	1) The geometric weight is small when the size difference between the two boxes is too large, \ie $\Delta w$ or $ \Delta h$ is close to 0 or too large.   
	2) The geometric weight tends to be larger when two objects are close to each other, \eg $(\Delta x, \Delta y) = (0.5, 0.5)$, than when their distance is large, \eg $(\Delta x, \Delta y) = (2, 2)$.

	\subsection{Examples Results of Generated Captions}
\begin{figure}[!tbh] 
		\centering
		\includegraphics[width=0.95\textwidth]{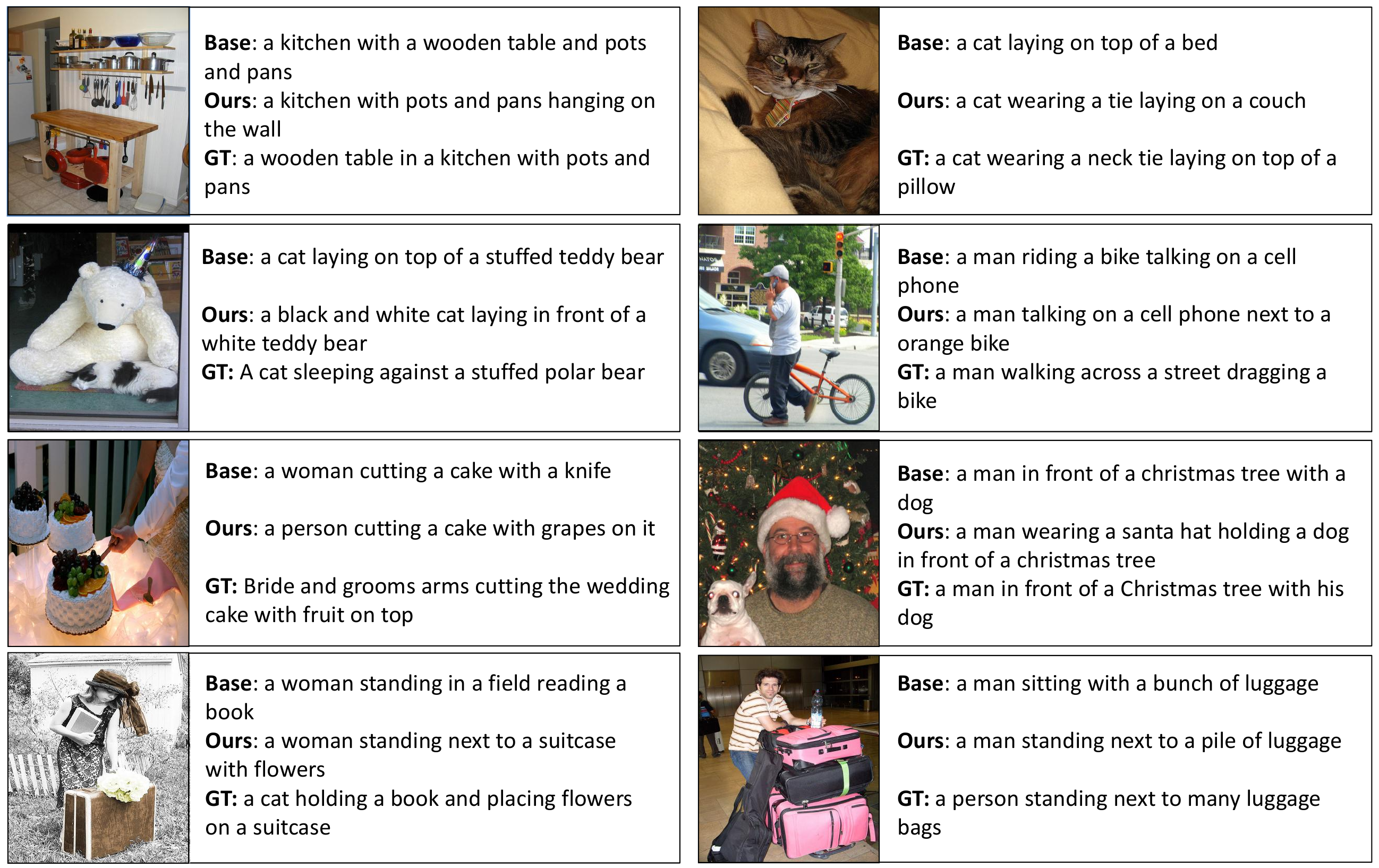}
		\caption{
			Example captions generated by the SAN baseline (denoted as `Base') and our G-SAN (denoted as `Ours') models. `GT' denotes one of the five ground-truth captions. 
			G-SAN shows an improvement over SAN in determining the relationships between objects. 
		}
		\label{fig:examples}
	\end{figure}

	To illustrate the advantages of the G-SAN relative to the SAN, 
	we present examples generated by the two models in Figure~\ref{fig:examples}.  
	The examples show that G-SAN is more good at determining the relationships between objects.  
	For example, in the first example, our G-SAN generates ``pots and pans \textit{hanging on} the wall", which successfully recognizes the relationship between 
	``pots and pans'' and ``wall", \ie ``hanging on".

\end{document}